\newif\ifarxiv
\newif\ifralfinal
\newif\ifconffinal
\ralfinalfalse \arxivtrue \conffinalfalse

\ifarxiv
\documentclass[letterpaper, 10 pt, conference]{ieeeconf}  %
\fi
\ifralfinal
\documentclass[letterpaper, 10 pt, journal, twoside]{IEEEtran}
\fi
\ifconffinal
\documentclass[letterpaper, 10 pt, conference]{ieeeconf} %
\fi

\IEEEoverridecommandlockouts

\usepackage{graphics} %
\usepackage{epsfig} %
\usepackage{times} %
\usepackage{amsmath} %
\usepackage{amssymb}  %
\usepackage{bbm}
\usepackage{booktabs}
\usepackage{adjustbox}
\usepackage{mathtools}
\usepackage{multirow}
\usepackage{units}
\usepackage{bm}
\usepackage{xcolor}
\let\labelindent\relax
\usepackage{enumitem}
\usepackage[ruled,vlined]{algorithm2e}
\usepackage{algorithmicx}
\usepackage{cite} 
\usepackage[caption=false, font=footnotesize]{subfig}
\usepackage{graphicx}
\usepackage{letltxmacro}
\LetLtxMacro{\originaleqref}{\eqref}
\renewcommand{\eqref}{Eq.~\originaleqref}

\SetCommentSty{mycommfont} %
\usepackage{microtype}

\newcommand{\I}[1]{\mathbb{I}_{#1}}

\ifarxiv
\usepackage{fancyhdr}
\fi
\makeatletter
\let\NAT@parse\undefined
\makeatother
\usepackage{hyperref}  %

\begin{document}
\title{
\ifarxiv\LARGE \bf\fi
Locking On: Leveraging Dynamic Vehicle-Imposed Motion Constraints to Improve Visual Localization
}
\author{
\ifarxiv
Stephen Hausler\authorrefmark{2}, Sourav Garg\authorrefmark{2}, Punarjay Chakravarty\authorrefmark{4}, \\ Shubham Shrivastava\authorrefmark{3}, Ankit Vora\authorrefmark{3} and Michael Milford$^*$\authorrefmark{2}%
\fi
\ifralfinal
Stephen Hausler$^{1}$, Sourav Garg$^{1}$, Punarjay Chakravarty$^{2}$, \\ Shubham Shrivastava$^{2}$, Ankit Vora$^{2}$ and Michael Milford$^{1}$%
\thanks{Manuscript received: February 24, 2022; Revised May 22, 2022; Accepted June 20, 2022.}%
\thanks{This paper was recommended for publication by Editor Sven Behnke upon evaluation of the Associate Editor and Reviewers' comments. 
This research was partially supported by funding from Ford Motor Corporation and NVIDIA, from ARC Laureate Fellowship FL210100156 to MM, and by the QUT Centre for Robotics.}
\fi
\thanks{\ifarxiv\authorrefmark{2}\fi 
\ifralfinal$^1$\fi The authors are with the QUT Centre for Robotics, School of Electrical Engineering and Robotics at the Queensland University of Technology.}%
\thanks{$^*$E-mail: \emph{firstname}.\emph{lastname}@qut.edu.au.}%
\thanks{\ifarxiv\authorrefmark{3}\fi
\ifralfinal$^2$\fi The authors are with the Ford Motor Corporation.}%
\thanks{\ifarxiv\authorrefmark{4}\fi
\ifralfinal$^3$\fi This work was conducted while the author was at the Ford Motor Corporation.}
\ifarxiv
\thanks{This research was supported by the Ford-QUT Alliance, NVIDIA, the QUT Centre for Robotics and ARC Laureate Fellowship FL210100156.}%
\fi
\ifralfinal
\thanks{Digital Object Identifier (DOI): see top of this page.} %
\fi
}
\bstctlcite{IEEEexample:BSTcontrol}

\ifralfinal
\markboth{IEEE Robotics and Automation Letters. Preprint Version. Accepted June, 2022}
{Hausler \MakeLowercase{\textit{et al.}}: Improving Worst Case Visual Localization}
\fi

\maketitle
\ifarxiv
\thispagestyle{fancy}
\pagestyle{plain}
\fi

\begin{abstract}

Most 6-DoF localization and SLAM systems use static landmarks but ignore dynamic objects because they cannot be usefully incorporated into a typical pipeline. Where dynamic objects have been incorporated, typical approaches have attempted relatively sophisticated identification and localization of these objects, limiting their robustness or general utility. In this research, we propose a middle ground, demonstrated in the context of autonomous vehicles, using dynamic vehicles to provide limited pose constraint information in a 6-DoF frame-by-frame PnP-RANSAC localization pipeline. We refine initial pose estimates with a motion model and propose a method for calculating the predicted quality of future pose estimates, triggered based on whether or not the autonomous vehicle's motion is constrained by the relative frame-to-frame location of dynamic vehicles in the environment. Our approach detects and identifies suitable dynamic vehicles to define these pose constraints to modify a pose filter, resulting in improved recall across a range of localization tolerances from $0.25m$ to $5m$, compared to a state-of-the-art baseline single image PnP method and its vanilla pose filtering. Our constraint detection system is active for approximately $35\%$ of the time on the Ford AV dataset and localization is particularly improved when the constraint detection is active.

\end{abstract}

\ifralfinal
\begin{IEEEkeywords}
Localization; Autonomous Vehicle Navigation; Deep Learning Methods; Multi Camera System
\end{IEEEkeywords}
\fi
\section{Introduction}

\begin{figure} %
    \centering
    \includegraphics[width=0.9\columnwidth, trim=0cm 17.1cm 0cm 0cm,clip]{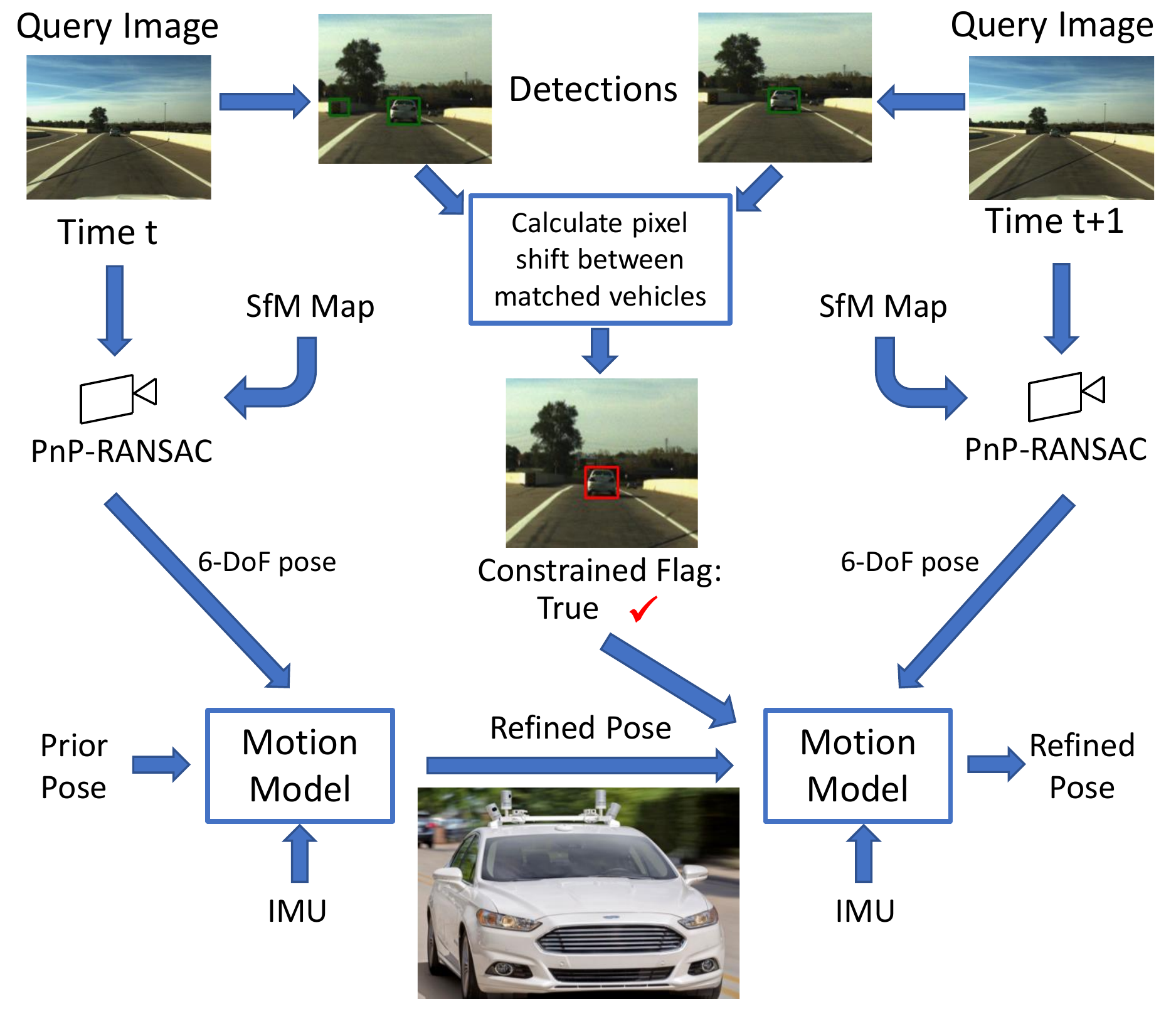}
    \caption{We propose an algorithm to first estimate whether detected dynamic vehicles are located at the same relative distance to the autonomous vehicle over time, and then use this to determine if the autonomous vehicle is likely following a \textit{constrained} motion path. Our pipeline uses vehicle detections and a motion model, to improve upon an initial pose estimate from PnP-RANSAC. Our constraint detector modifies our motion model to consider whether future pose estimates are consistent with the constraints imposed by the relative positioning of dynamic vehicles in the environment. Our work utilizes one of Ford's autonomous vehicles.}
    \vspace*{-0.2cm}
    \label{fig:pipeline_figure}
\end{figure}

For autonomous vehicles to operate safely and effectively, robust and accurate localization is critical, with positional accuracy requirements in the order of 10-20cm~\cite{reid2019localization}. Current solutions to this challenge in the industry commonly use a sensor fusion of GPS (Global Positioning System), INS (Inertial Navigation System) and LiDAR, using techniques such as Extended Kalman Filter~\cite{Wolcott17, vora2020aerial}. 6-DoF visual localization~\cite{Sattler2018,pion2020benchmarking} using cameras offers an alternative approach that can attain highly accurate localization in autonomous driving scenarios~\cite{Hausler2022}, and has the benefits of reducing reliance on expensive LiDAR sensors and providing additional localization redundancy. State-of-the-art feature-based visual localization pipelines provide localization estimates on a per-frame basis, and are prone to failures due to inaccuracies in 3D reconstruction or feature description and association.

In this work, we propose an approach for refining frame-to-frame pose estimates using a trigger-constrained motion model that considers the relative position of dynamic vehicles in the environment, whenever available. We propose to leverage the detection of dynamic vehicles, already typically performed in an AV autonomy stack for other purposes, as a limited but still beneficial source of information within a localization pipeline. To do so, we consider sequential filtering of 6-DoF poses obtained from single image, PnP-RANSAC based visual localization and improve the filter's pose prediction through the proposed motion constraints of dynamic vehicles. We re-use the local features from the localization pipeline and semantic segmentation masks from the perception pipeline of the AV to generate vehicle descriptors, which we use to compare dynamic vehicles across consecutive images in order to determine if their relative position is constant with respect to the autonomous ego-vehicle. If their relative position is constant over time, we make the approximation that the autonomous vehicle's velocity and heading are also constant and therefore constrained. Then, building upon an Extended Kalman Filter sensor fusion system, we use this external pose constraint information, whenever available, to modify the contribution of this pose estimate to the frame-by-frame localization state of the filter. 

We experimentally validate our proposed approach using 42 km of driving from the Ford AV dataset~\cite{agarwal2020ford}, divided into 247 segments of length 150m, across a variety of traffic conditions, road types (highway, suburban) and times of day.

\section{Related works}
In this section, we review recent research on 6-DoF visual localization including the use of semantic object detection, tracking and pose constraining in relation to localization.

\subsection{6-DoF Visual Localization}
6-DoF visual localization is the task of estimating absolute camera pose given a prior 3D map and a query image~\cite{Sattler2018}. This remains a challenging research problem especially for long-term and continuous operations of autonomous vehicles, where query images can undergo significant appearance and viewpoint variations as compared to the database images. In recent years, researchers have explored different approaches to address such issues, which include learning robust local features~\cite{DeTone18,Dusmanu2019CVPR,lee2021local,zhou2020da4ad,venator2021self}, planes~\cite{jeon2020learning}, matchers~\cite{sarlin2020superglue,clement2020learning} and global descriptors~\cite{pion2020benchmarking}, leveraging sequential information~\cite{stenborg2020using,lee2021local}, improving 2D-3D matching~\cite{song2021recalling}, and combining semantics and geometry~\cite{larsson2019fine,stenborg2018long,huang2021vs,han2020scene}. Learning-based methods have also been developed for scene coordinate regression~\cite{shotton2013scene,li2020hierarchical,huang2021vs}, robust pose estimation~\cite{brachmann2017dsac} and direct image/feature alignment~\cite{Stumberg20,sarlin2021back,lindenberger2021pixel}. In this work, we use a feature-based visual localization pipeline as in~\cite{Sattler2018}, implemented using Kapture~\cite{humenberger2020kapture} and HLoc~\cite{Sarlin19}, and improve localization using dynamic vehicles in the scene -- an aspect complementary to the above-listed recent innovations in this field.

\subsection{Localization and Semantic Scene Understanding}
Several methods exist that incorporate semantic segmentation and object detection into visual place recognition and 6-DoF metric localization pipelines, as surveyed in~\cite{garg2020semantics}. This includes using Faster-RCNN~\cite{ren2015faster} for constructing object graphs~\cite{han2018learning} or object matching~\cite{zhang2019coarse}; segmenting specific objects/entities such as buildings~\cite{armagan2017learning}, lanes~\cite{schreiber2013laneloc} and skyline~\cite{saurer2016image} for improved recognition; or pre-selecting multiple object classes~\cite{naseer2017semantics,garg2018lost,Gawel2018}. In particular for 6-Dof localization, semantic label consistency between 3D points and their projections on the query image was employed within a particle filter~\cite{stenborg2018long} or a P3P-RANSAC loop~\cite{toft2018semantic,larsson2019fine} for improving pose estimation. \cite{schonberger2018semantic} used semantic scene completion to learn 3D descriptors and align point clouds. \cite{radwan2018vlocnet++} used multi-task learning to jointly estimate ego-motion, global pose and per-pixel semantic labels. In all these approaches, either it is assumed that dynamic objects are not useful during localization and thus removed, or such objects do not form a part of the 3D map during SfM reconstruction. In this work, we instead explore the use of such dynamic objects to improve metric localization.

\subsection{Localization and Object Tracking}
Object tracking has applications in various related fields such as augmented reality~\cite{park2008multiple}, action recognition~\cite{wang2016action}, and path planning for autonomous vehicles~\cite{leon2021review}.
Researchers have also combined (ego-vehicle) localization and dynamic object tracking, where either the former informs the latter~\cite{vu2007online,phillips2021deep}; the latter informs the former~\cite{migliore2009use,gohring2006multi}; or both are jointly modeled for the mapping task~\cite{wang2007simultaneous,liu2021switching}. 
\cite{migliore2009use} tracks moving features but only uses static features for the localization part. \cite{gohring2006multi} is able to localize a robot by tracking a moving object but only in a multi-robot setting. We propose a new and different approach: using the dynamic vehicle tracking signal directly to improve ego-vehicle localization in the context of autonomous driving.

\subsection{Constraints-based Camera Pose Estimation}
There exist several methods that constrain camera pose estimation using additional information. This includes distance constraints between semantic components and 3D point cloud~\cite{liang2022semloc}, leveraging direction of gravity and camera height~\cite{svarm2014accurate}, combined point and line constraints~\cite{ramalingam2011pose}, using geo-referenced traffic signs~\cite{qu2015vehicle}, and constraints based on environment-object distances~\cite{huang2018cooperative}, and mirror reflections~\cite{takahashi2016mirror}. However, the exact methodology varies widely in these works depending on the context of application and the assumptions of additional information. Distinct from these approaches and considering autonomous driving applications, we propose to use velocity constraints based on the motion of dynamic vehicles relative to the ego-vehicle to improve ego-vehicle localization.

\section{Methodology}

This section proceeds as follows: we begin by describing the key ideas of our approach, followed by description of the concepts used in our system. Our contributions are then described, providing an algorithm for improving pose filtering using motion constraints from dynamic vehicles.

In this work, we consider whether the localization performance of an autonomous vehicle can be improved by considering the motion of external dynamic vehicles moving ahead of the ego-vehicle, by using these vehicles as a pose constraint. We propose the following simple constraint definition: if the 2D pixel location of a dynamic vehicle does not vary between two consecutive images, then we assume that the relative position of the ego vehicle and the dynamic vehicle is constant. Furthermore, we can then approximate the velocity and heading to also be constant over this duration. This enables us to create a conditional algorithm, to detect whether the position of the dynamic vehicle in the image plane is stationary and adjust the non-holonomic motion constraints based on this condition; in the absence of constraints, the pose filter operates as is. This constraint definition, while simple, has the advantage of requiring very little of the complicated detection and object pose estimation and tracking machinery of other approaches: and we are able to evaluate whether it works in practice in real world AV related experiments. 

\subsection{SfM Visual Localization Pipeline}
Our proposed method for relative motion-constrained filtering is agnostic to the exact source of single image 6-DoF poses, thus any of the existing state-of-the-art single image visual localization approaches could be used. We chose local feature-based visual localization pipeline HLoc~\cite{Sarlin19} with R2D2~\cite{revaud2019r2d2} local features, as we re-purpose these features for motion constraint detection (see Section~\ref{subsec:constraint}). HLoc uses PyCOLMAP~\cite{schoenberger2016sfm} for SfM reconstruction and PnP-RANSAC for estimating 3D pose for a single query image, which we use in a sequential filter as described next.

\subsection{EKF Pose Refinement with IMU - Baseline System}

We model the AV using the bicycle model and implement a 6-DoF Error State Extended Kalman Filter~\cite{Kalman1960, markovic2021error, mourikis2007multi} (hereafter referred to as the EKF) to filter the frame-to-frame estimated poses with the addition of a motion model. We also assume that the AV platform has an IMU (Inertial Measurement Unit), and we integrate the IMU readings into the EKF. The IMU data comprises linear acceleration ($I_a$) and angular velocity ($I_v$), with the following state vectors:
\begin{equation}
    i_a = [a_x \quad a_y \quad a_z], \quad
    i_v = [\omega_x \quad \omega_w \quad \omega_z]
\end{equation}
The prediction updates of the EKF are:
\begin{equation}
    p_k = p_{k-1} + \delta v_{k-1} + \frac{1}{2}\delta^2 R_{k-1}i_a
\end{equation}
\begin{equation}
    v_k = v_{k-1} + \delta R_{k-1}i_a
\end{equation}
\begin{equation}
    q_k = q\{\delta I_v\} \otimes q_{k-1}
\end{equation}
where $\delta$ is the time difference between consecutive iterations of the filter; $R_{k-1}$ is the rotation matrix for the current orientation of the vehicle; and $p_k, v_k, q_k$ are the three state variables of our filter representing the $x,y,z$ position, velocity and the quaternion respectively. The filter operates with respect to image timestamps and the IMU vector is calculated by numerically integrating the raw IMU data at the timestamp that the image is captured. Following~\cite{Mourikis2007}, we use the symbol~$\otimes$ to denote quaternion multiplication and $q\{\}$ to denote the transform from rotation vector to quaternion representation.

We compute the process Jacobian as follows:
\begin{equation}
    A = \begin{bmatrix}
    a_x \\ a_y \\ a_z
    \end{bmatrix}_\times
    F_x = \begin{bmatrix}
    \I{3} & \I{3}\delta & 0_{3\times3}\\
    0_{3\times3} & \I{3} & -R_{k-1}A\delta\\
    0_{3\times3} & 0_{3\times3} & R^T\{\I{}v\delta\}\\ %
    \end{bmatrix}
\end{equation}
where $R\{x\}$ denotes the rotation matrix representation of an angle-axis representation $x$ and $[~]_\times$ denotes a vector as a skew-symmetric matrix. The process noise Jacobian is defined through process variance $v_p$:
\begin{equation}
    Q = \begin{bmatrix} \I{6}v_p\delta^2 \end{bmatrix}
\end{equation}
with the covariance $C_k$ and Kalman gain $G$ updated as:
\begin{equation}
    C_k = F_xC_{k-1}F_x^T + F_wQF_w^T
\end{equation}
\begin{equation}
    G =C_kH^T\left(HC_kH^T + v_m\I{6}\right)^{-1}
\end{equation}
where $F_w$ and $H$ denote Jacobian of the motion model noise and measurement model:
\begin{equation}
    F_w = \begin{bmatrix}
    0_{3\times3} & 0_{3\times3}\\
    \I{3} & 0_{3\times3}\\
    0_{3\times3} & \I{3}
    \end{bmatrix}
\end{equation}
From here we will define the state variables $p_k, v_k, q_k$ as $x$ for simplicity.
Now we will define the measurement update equations.
First, the state error $\delta x$ is calculated - this is the difference between our estimated state variables $x$ and the incoming measured state variables $y$.
\begin{equation}
    \delta x = G\left(y-x\right)
\end{equation}
Then we calculate our corrected state variables by updating $x$, with the updated x denoted as $\hat{x}$:
\begin{equation}
    \hat{x} = x + \delta x
\end{equation}
Noting that in practise, this may be a simple linear addition for positional components and quaternion multiplication for rotational components.
Then we update the filter covariances $C_k$.
\begin{equation}
    \hat{C_k} = \left(\I{9} - GH\right)C_k
\end{equation}
Noting that for numerical stability the Joseph form may be used instead.
Then we process the ESKF reset, which further updates $\hat{C_k}$:
\begin{equation}
    \hat{C_k} = J\hat{C_k}J
\end{equation}
Where $J$ is a Jacobian defined as:
\begin{equation}
    J = \begin{bmatrix}
    \I{3} & 0_{3\times3} & 0_{3\times3}\\
    0_{3\times3} & \I{3} & 0_{3\times3}\\
    0_{3\times3} & 0_{3\times3} & \I{3}-0.5\delta\theta
    \end{bmatrix}
\end{equation}
Where $\delta\theta$ is the angular components of the state error $\delta x$.

Using single image 6-DoF PnP pose estimates as measurements, we follow the standard approach~\cite{sola2017quaternion, markovic2021error} to correct $p_k, v_k, q_k, C_k$, where the posterior estimates of state variables are obtained by adding the prior estimates and the gain-weighted residual (i.e., the difference between the prior estimate and the current measurement). In our experimental results, this setup is referred to as the baseline EKF.

\subsection{Dynamic Vehicle Constrained Pose Filtering}

We now consider the motion model with the knowledge that the motion of the AV will be constrained to a constant velocity and heading. Whenever a constraint is detected (as described in subsequent sections), for a given measured pose $M_k$, we can expect the subsequent measured pose $M_{k+1}$ to be located at the position of the current measured pose plus a distance vector ($M_k + \delta V_k$). Therefore, we can adjust how much to trust our next measurement based on how proximal the subsequent measurement is relative to where we expect the measurement to be.
This adjustment is done through the measurement noise variance depending on the distance between our estimated measurement and our actual measurement. It can also be estimated that $M_{k+1}$ is likely to be proximal to $P_{k+1}$; however, if the EKF experiences drift this can cause a scenario where future measurements are insufficiently trusted. $V_k$ can also drift, but the magnitude of drift will be less since velocity is the derivative of position.

In practice, we observe that deviations in individual pose measurements are detrimental to the filtered pose estimate. To mitigate this problem, we make the design decision to use a Radial Basis Function (RBF) Kernel in order to penalize large deviations between expected and actual pose measurements. We also observed that using a non-linear RBF kernel provides more accurate localization compared to using the Euclidean distance metric.

We now define our dynamic measurement variance $v'_m$ by applying a Radial Basis Kernel to the axes displacements (translation component only) between the actual measurement, $M_{k+1}$ to the predicted measurement, $\bar{M}_{k+1}$:
\begin{equation}
    \bar{M}_{k+1} = M_k + \delta V_k
\end{equation}
\begin{equation}
    Kx = \exp \left( \frac{-||M_{k+1}x - \bar{M}_{k+1}x||^2}{2\sigma_x^2} \right)
\end{equation}
\begin{equation}
    Ky = \exp \left( \frac{-||M_{k+1}y - \bar{M}_{k+1}y||^2}{2\sigma_y^2} \right)
\end{equation}
\begin{equation}
    Kz = \exp \left( \frac{-||M_{k+1}z - \bar{M}_{k+1}z||^2}{2\sigma_z^2} \right)
\end{equation}
where $x,y,z$ are the three translational axes of the SfM coordinate frame. $\sigma_x, \sigma_y, \sigma_z$ denote the bandwidth of the kernels. If the AV is not constrained, we can still expect the motion to be predictable knowing that the velocity and heading of the AV will not change instantly. When the constraint detection is true, we decrease the values of $\sigma_x$ and $\sigma_y$ by a factor $\alpha$ ($\sigma_z$ will always be constrained to the road surface) to decrease the RBF error which in turn increases the dynamic measurement variance, defined as follows:
\begin{align}
    v'_m = v_m + \left(\frac{1}{Kx} - 1 \right) + \left(\frac{1}{Ky} - 1 \right)
    + \left(\frac{1}{Kz} - 1 \right) 
\end{align}
When the AV is constrained, the smaller value of $\sigma$ results in a smaller bandwidth of the RBF, which then causes deviations in the expected measurement to result in a smaller value of $K$. A smaller value of $K$ then results in a larger value of $v_m$, which means that the filter will be adjusted by a smaller magnitude from the new measurement. Intuitively, when the constraint detection is active the filter will only consider new measurements that are consistent with the current motion pattern of the AV.

\subsection{Vehicle Detection}

Our constraint algorithm first requires the detection of vehicles in each image in the form of pixel-level masks.
We use dense semantic segmentation network, Panoptic-Deeplab-v3~\cite{cheng2020panoptic}, to obtain vehicle-labelled pixels. In order to decouple the effect of perception noise (leading to false positive detections) from localization performance of the proposed method, we combine the semantic segmentation output with an object detector network, YOLO-P~\cite{wu2022yolop} to emulate a high-precision vehicle detection system of an AV.
Using YOLO-P, we extract bounding boxes for all objects observed in the image that are classified as vehicles. We then mask multiply the segmentation result from Panoptic-Deeplab (for the car, truck and bus classes) with the YOLO-P bounding box area for each detected vehicle. This returns a per-pixel segmentation only considering the vehicles themselves, with no background pixels. We also check the size of each per-pixel vehicle mask and remove masks under 0.04\% pixel area - this is to remove very small detections which do not have sufficient size for accurate calculation of the constraint detection.

\subsection{Constraint Detection}
\label{subsec:constraint}
In this final subsection, we describe our approach for calculating whether or not the AV is in a constrained state, based on our detection and analysis of dynamic vehicles in the environment. Our objective is to identify if any of the dynamic vehicles in the scene are (apparently) stationary with respect to the AV between two consecutive images, since a vehicle that is apparently stationary in the image plane is moving at approximately the same heading and velocity as the AV. We make the assumption in this work that a vehicle can approximately be considered a planar surface. 

\paragraph*{Vehicle Appearance Representation} For each pixel in each detected vehicle, we obtain a local feature from the 3D tensor of R2D2's descriptor head, precomputed during the single image PnP localization.
We categorize each local feature by the vehicle it was produced from, such that we have a set of local features per vehicle (denoted as $L_n$, where $n$ is the index of the current vehicle). We also calculate an overall per vehicle descriptor by average pooling the set of local descriptors describing the vehicle; this produces a 128-dimensional descriptor per vehicle. This procedure is repeated for every query frame.

\paragraph*{Vehicle Recognition and Tracking} For the constraint decision process, we match local vehicle descriptors between subsequent images (denoted as $IM_1$ and $IM_2$). First, the per vehicle descriptors are used to perform data association between the sets of vehicles in each image. For each successfully matched pair of vehicles, we have two sets of local descriptors $L_1^n$ and $L_2^m$, where $n$ and $m$ denote the vehicle index within images $IM_1$ and $IM_2$. We then use mutual nearest neighbours to find local feature correspondences between sets of descriptors $L_1^n$ and $L_2^m$. 

\paragraph*{Locking on Dynamic Vehicle} We find the per-pixel keypoint shift between the set of local matches and calculate the mean pixel shift $\mu_p$. If $\mu_p$ is less than a threshold $\tau$, we consider this particular pair of vehicle instances (which is the same vehicle but temporally shifted) to be constrained with respect to the autonomous vehicle. This process is repeated for each matched pair of vehicles across the two images. If at least one pair of vehicles is found to be constrained, then we consider the AV to be constrained for the current frame $IM_2$.

\section{Experimental Setup}

We evaluate our proposed system in a varied range of real-world autonomous driving conditions, as described in this section.

\subsection{Datasets}
To evaluate our proposed technique, we use the Ford AV dataset~\cite{agarwal2020ford}. The dataset is split up into a number of logs, which consist of different driving routes around Michigan at different times of the year. For all experiments, we use the image data from the front-left forward-facing camera, sampled every 1.5 meters. Ground truth information is provided by the dataset authors, using 6-DoF lidar pose-graph SLAM with full bundle adjustment. We use \emph{2017-08-04} as the mapping traverse and \emph{2017-10-26} as the query traverse. We evaluate using Logs 1, 2 and 4, corresponding to the conditions of Highway (13km), Airport/Roadwork (24km) and Uni campus (5 km).

\subsection{Implementation Details}

\subsubsection{Dataset Preprocessing}

Given a log of data, we split the log up into a sequence of approximately 1 km slices, and produce separate SfM reconstructions for each slice. This is standard practice as a complete 3D reconstruction of the whole log is prone to errors~\cite{Hausler2022, Sattler2018}. We also remove any sections of the log where the mapping and query traverses do not overlap.

\subsubsection{EKF Design}

Our EKF is initialized at the first PnP-RANSAC pose estimate, with $P_0$ equal to the translation component of the first measurement and $Q_0$ equal to the rotation component. $V_0$ is calculated using the first 10 measured poses and the timestamp difference between consecutive images. The magnitude of the velocity is computed using translation components, while the direction of the velocity vector is calculating using the rotation components. Since our proposed modification of measurement variance is based on the baseline $v_m$, we perform sensitivity analysis using Log 5 of the Ford dataset to set $v_m = 0.005$, as per Table~\ref{tab:vm}, where Recall@0.5m can be observed to saturate for lower values of $v_m$. The process variance $v_p$ is always fixed to 0.5. The RBF Kernel is configured with $\sigma_x = \sigma_y = 2.6$, $\sigma_z = 2.1$ and $\alpha = 2.0$. Similar to $v_m$, these parameters were set through Log5 calibration subset. Our constraint detection threshold $\tau$ is set as a proportion of the detection bounding box size (to account for vehicles at different depths), specifically: $\tau = \frac{\sqrt{A_b}}{70}$, where $A_b$ defines the area of the current bounding box. We utilize the IMU data provided in the Ford AV dataset, which has an Applanix POS LV inertial measurement unit~\cite{agarwal2020ford}.

\begin{table}
    \centering
    \caption{Sensitivity Analysis of Measurement Variance}
    \begin{tabular}{ccccccc}
    \toprule
         $v_m$ & 0.001 & 0.005 & 0.01 & 0.05 & 0.1 & 0.5   \\
    \midrule
         R@0.5m & 0.982 & \textbf{0.983} & 0.981 & 0.970 & 0.945 & 0.862  \\
    \bottomrule
    \end{tabular}
    \label{tab:vm}
\end{table}

\subsection{Evaluation}

Our algorithm is designed to filter and predict pose estimates by tracking dynamic vehicles and then constraining the frame-by-frame pose estimates over a traverse. To thoroughly evaluate the performance of the algorithm, we split each 1 km slice into 150m long segments, in order to verify the performance of our system with different initial starting conditions. In total, we test using 247 segments and compile results based on the cumulative localization performance across all segments.

We follow the standard evaluation as proposed in~\cite{Sattler2018} for 6-DoF localization, where three different bins of translation and rotation errors are used with a varying degree of localization precision: 0.25 meters and 2\textdegree,  0.50 meters and 5\textdegree, and 5 meters and 10\textdegree. Furthermore, we record the localization error at the end of each 100m segment along with the `worst-case' localization error, that is, the maximum individual pose error in a segment (only considering the translation component)~\cite{Hausler2022}.

\section{Results}\label{sec:results}

In the first subsection, we detail the localization accuracy of our approach under different environmental conditions. Then, we present qualitative examples to show the performance of the algorithm in a variety of different subsections of the dataset.

\subsection{Quantitative Performance}
We evaluate our proposed approach on three different traverses of the FordAV dataset, covering different road conditions over a total of 42km of driving. We show results on three key settings: a) \textit{PnP}: single image localization based on R2D2 and PnP-RANSAC~\cite{Sarlin19,revaud2019r2d2,humenberger2020kapture}, b) \textit{EKF}, which uses these PnP estimates and IMU to predict poses, and c) \textit{Ours}, where this EKF is updated if the dynamic vehicle constraint is triggered. A summary of our results is shown in Table~\ref{tab:endmaxerr} and~\ref{tab:recall}.

As per previous work~\cite{Hausler2022}, we first examine maximum translation errors within each segment, for our approach and baselines. We report maximum error (\textit{MaxErr} in Table~\ref{tab:endmaxerr}) with its mean and median over all segments of a given log. As shown in Table~\ref{tab:endmaxerr}, for three different environmental conditions respectively, we find that our pose refinement reduces the mean \textit{MaxErr} from $8.34m$, $5.21m$ and $11.09m$ with PnP-only down to $3.05m$, $2.8m$ and $1.43m$, while simultaneously reducing the median error. Since an EKF by design is suited to filtering poor individual estimates, we observe that EKF-only reduces the maximum errors to a mean of $3.69m$, $3.35m$ and $4.19m$; however, our constraint-triggered dynamic measurement variance further aids in removing the individual poor PnP estimates.

In Table~\ref{tab:recall}, we compare recall performance at different localization precision levels, as defined in~\cite{Sattler2018}. 
Our proposed algorithm attains $60.8\%$, $63.9\%$ and $88.6\%$ recall at $0.25m/2^{\circ}$, compared to $58.2\%$, $63.5\%$ and $88.5\%$ with PnP-only. With EKF-only, performance either increases only marginally ($+0.6\%$ on highway) or decreases ($-0.6\%$ on airport and $-1.0\%$ on campus logs). We observe that the baseline filter is sensitive to measurement errors and any deviations in the filter state as a result of erroneous measurements result in reduced localization performance at the $0.25m$ tolerance. The addition of a dynamic measurement variance with motion constraints successfully removes this limitation. A similar trend exists at the $0.5m$ tolerance. We notice that our technique also achieves the highest recall at the $5m$ tolerance, except on the highway condition. We observe that filtering can still fail to prevent inaccurate localization if a consistent set of poor measurements is encountered, such as in the case of a pose drift over time.

We also consider some alternative metrics, to understand the performance of the algorithm over the course of each segment. We measure the translational pose error at the end of each 100 image segment, and calculate the mean and median errors for all 73 segments (shown in Table~\ref{tab:endmaxerr}). We observe that both the mean and median errors 
are reduced with our approach versus baseline methods at the end of each segment. A discrepancy exists during the airport condition where the mean end error is larger - an edge case failure mode is where the constraint detection activates when it should not (such as when the vehicle is turning a corner), which then causes the filtered pose to deviate from the ground truth.

\begin{table}[]
\caption{Mean/Median over End and Max Error of segments for different environmental conditions.}
\label{tab:endmaxerr}
\begin{adjustbox}{width=0.5\textwidth}
\begin{tabular}{@{}lcccccc@{}}
\toprule
              & \multicolumn{2}{c}{\textbf{Highway}}    & \multicolumn{2}{c}{\textbf{Airport/Roadwork}} & \multicolumn{2}{c}{\textbf{Uni campus}} \\ 
\cmidrule(lr){2-3} \cmidrule(lr){4-5} \cmidrule(lr){6-7}

              & \textit{End Err}   & \textit{Max Err}   & \textit{End Err}      & \textit{Max Err}      & \textit{End Err}   & \textit{Max Err}            \\
              \midrule
PnP           & 0.90/0.20          & 8.34/1.28          & 1.00/\textbf{0.15}             & 5.21/1.02             & 0.15/\textbf{0.06}          & 11.09/0.40         \\
EKF           & 0.78/0.19          & 3.69/0.75          & \textbf{0.65}/0.18             & 3.35/0.59             & 0.13/\textbf{0.06}          & 4.19/\textbf{0.37}          \\
\textbf{Ours} & \textbf{0.71/0.17} & \textbf{3.05/0.62} & 1.33/0.17            & \textbf{2.80/0.56}     & \textbf{0.12/0.06} & \textbf{1.43}/0.41 \\
\bottomrule
\end{tabular}
\end{adjustbox}
\end{table}

\begin{table}[]
\caption{Recall per localization precision for different environmental conditions.}
\label{tab:recall}
\begin{adjustbox}{width=0.5\textwidth}
\begin{tabular}{lccccccccc}
\toprule
\textbf{} & \multicolumn{3}{c}{\textbf{Highway}}          & \multicolumn{3}{c}{\textbf{Airport/Roadwork}} & \multicolumn{3}{c}{\textbf{Uni campus}}       \\ 
\cmidrule(lr){2-4} \cmidrule(lr){5-7} \cmidrule(lr){8-10}

         \small{R@m/°} & \scriptsize{.25/2}       & \scriptsize{.5/5}        & \scriptsize{5/10}        & \scriptsize{.25/2}       & \scriptsize{.5/5}       & \scriptsize{5/10}         & \scriptsize{.25/2}       & \scriptsize{.5/5}        & \scriptsize{5/10}        \\
\midrule
PnP       & 58.2          & 77.2          & 98.1          & 63.5          & 81.3         & 97.6           & 88.5 & 95.3          & 99.1          \\
EKF       & 58.8          & 77.6          & \textbf{98.4} & 62.9          & 81.3         & \textbf{97.8}  & 87.5          & 95.2          & 99.2            \\
\textbf{Ours}      & \textbf{60.8} & \textbf{79.9} & 98          & \textbf{63.9}   & \textbf{83.5}  & \textbf{97.8}           & \textbf{88.6} & \textbf{96} & \textbf{99.7} \\ \bottomrule
\end{tabular}
\end{adjustbox}
\end{table}

\label{table:new}

\begin{figure*}  %
    \centering
    \includegraphics[width=0.90\linewidth, trim=0cm 25cm 0cm 0cm,clip]{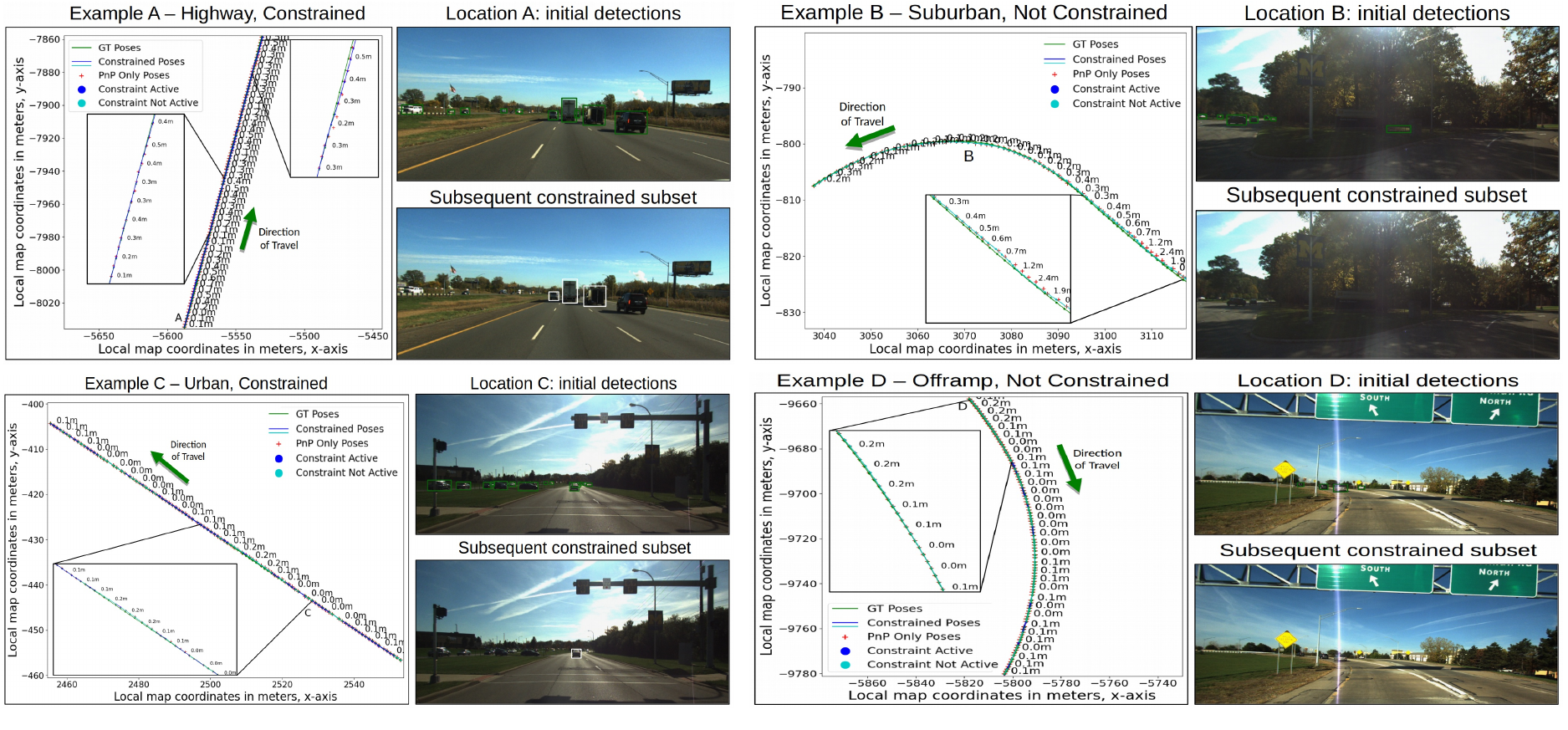}
    \caption{Qualitative results for four examples from the dataset. In each example, we show a metric coordinate map for a segment and the numbers on the map denote the pose error from the constrained pose (our proposed approach) to the ground truth. On the right-hand side of each example, images are shown from the location marked with a letter on the metric map. Top image: initial detections from YOLOP, marked in green. Bottom image: the subset of detections that are deemed to be constrained relative to the AV, marked in white.}
    \label{fig:example}
    \vspace*{-0.2cm}
\end{figure*}

\subsection{Analysis of Constraint Detection Performance}

To further analyse the impact of our constraint detection system, we analysed the localization performance split across different sections of each traverse depending on whether or not constraint detection was active in that section, as shown in Table~\ref{tab:constaintablation}.
Across the three traverses, the improvement of \emph{Ours} over the baseline systems (\emph{PnP}, \emph{EKF}) is greater when the constraint is active. For example, on the airport traverse with the constraint active, our proposed system achieves a recall of $80.6\%$ recall at $0.5m/5^{\circ}$, while PnP-only and EKF-only achieves $77.1\%$ and $77.5\%$ respectively. When we consider the sections of the traverse where the constraint detection was inactive, we find that our proposed system achieves a smaller margin of improvement over the baselines. Without constraints, \emph{Ours} achieves a recall of $85.6\%$ at $0.5m/5^{\circ}$, versus $84.4\%$ and $84.1\%$ for PnP-only and EKF-only. 

It is also important to note the frequency at which the constraint detection is activating; the constraint detection is active for $56.8\%$ of the distance travelled on the highway traverse, $41.2\%$ on the airport traverse and $8.0\%$ on the campus traverse. The large discrepancy is due to the varied nature of these traverses, since the highway and airport routes contain many sections of long, straight roads and the campus route contains many curves and intersections.

\begin{table}[]
\caption{Recall per localization precision when/when-not constrained.}
\label{tab:constaintablation}
\begin{adjustbox}{width=0.5\textwidth}
\begin{tabular}{lcccccccccl}
\toprule
\textbf{} & \multicolumn{3}{c}{\textbf{Highway}}          & \multicolumn{3}{c}{\textbf{Airport/Roadwork}} & \multicolumn{3}{c}{\textbf{Uni campus}}       \\ 
\cmidrule(lr){2-4} \cmidrule(lr){5-7} \cmidrule(lr){8-10}

         \textbf{Constraint Active} & \scriptsize{.25/2}       & \scriptsize{.5/5}        & \scriptsize{5/10}        & \scriptsize{.25/2}       & \scriptsize{.5/5}       & \scriptsize{5/10}         & \scriptsize{.25/2}       & \scriptsize{.5/5}        & \scriptsize{5/10} & \scriptsize{R@m/°}       \\
\midrule
PnP       & 50.3          & 72.1          & 97.4          & 52.6          & 77.1         & 98.3           & 78.9 & 92.7          & 98.5          \\
EKF       & 50.9          & 72.7          & \textbf{97.5} & 52.9          & 77.5         & \textbf{98.7}  & 78.9          & 95.8          & 98.9            \\
\textbf{Ours}      & \textbf{53.3} & \textbf{75.1} & 96.9          & \textbf{54.7}   & \textbf{80.6}  & 98.4           & \textbf{80.8} & \textbf{96.6} & \textbf{100} \\
\textbf{Ours - Delta}      & \textbf{+2.4} & \textbf{+2.4} & -0.6          & \textbf{+1.8}   & \textbf{+3.1}  & -0.3           & \textbf{+1.9} & \textbf{+0.8} & \textbf{+1.1} \\
\midrule
    \multicolumn{11}{l}{\textbf{Constraint Inactive} (note: absolute performance varies due to a different selection of images)} \\
\midrule
PnP       & 68.6          & 83.9          & 99.1          & \textbf{71.4}         & 84.4         & 97.1           & \textbf{89.4} & 95.6          & 99.2          \\
EKF       & 69.2          & 84.1          & \textbf{99.5} & 70.1          & 84.1         & 97.1  & 88.3          & 95.1          & 99.2            \\
\textbf{Ours}      & \textbf{70.7} & \textbf{86.2} & 99.4          & 70.5   & \textbf{85.6}  & \textbf{97.4}           & \textbf{89.4} & \textbf{96} & \textbf{99.7} \\
\textbf{Ours - Delta}      & \textbf{+1.5} & \textbf{+2.1} & -0.1          & -0.9   & \textbf{+1.2}  & \textbf{+0.3}           & 0 & \textbf{+0.4} & \textbf{+0.5} \\\bottomrule    
\end{tabular}
\end{adjustbox}
\end{table}
\label{table:two}

\subsection{Qualitative Examples}

In Figure~\ref{fig:example}, we showcase four examples of our algorithm in action. In each example, we show a birds-eye view of both the estimated poses from PnP and our constrained filter along with ground truth poses, for all the images from a segment. We also provide example images from within that subset, showing both the vehicle detections produced by YOLOP and the final detected constrained vehicles. 

In example A, the constraint detector is constantly active, because the autonomous vehicle is travelling at a constant speed on the highway and in a straight line. The detector has locked onto three vehicles which are travelling at the same velocity. In the zoomed insets, it can be observed that the constrained EKF system responds slowly to measurements that are inconsistent with the existing velocity state of the system (see the red PnP only poses).

Example B is a situation where the constraint detector is inactive, since the autonomous vehicle is travelling around a corner. In example C, we show a constrained example in an urban area. In this example, the constraint detector is not constantly active during the segment. This is due to slight changes in the relative velocity between the two vehicles in this urban area, noting that there is only a single vehicle that is valid for locking on to in this example.

Example D shows another not-constrained case, where the autonomous vehicle is travelling around a shallow curve. This is a partial failure case, since the constraint detector is briefly activating during the curve. Because the curve is very wide, and the velocity of the vehicles are constant, this is an edge case that sits between constrained and unconstrained motion profiles.

\begin{table}
    \centering
    \caption{Compute time for different components of our pipeline.}
    \begin{tabular}{ccccc}
    \toprule
         \textbf{Method} & PnP & Perception & Constraint Det. & Filter  \\
    \midrule
         \textbf{Time} (ms) & 1890 & 798/122 & 43 & 8.6 \\
    \bottomrule
    \end{tabular}
    \label{tab:compute}
\end{table}
\subsection{Computation Analysis}

Table~\ref{tab:compute} shows computation time for different components of our pipeline (on a per image basis), using a desktop computer with a 4.20GHz 8-Core i7-7700K CPU and a Nvidia 1080Ti GPU. \textit{PnP} refers to the single image-based 6-DoF pose estimation method (HLoc); \textit{Perception} refers to single image semantic segmentation and object detection (i.e., Panoptic DeepLabv3 / YOLOP); \textit{Constraint Det.} refers to our method in Section~\ref{subsec:constraint}, and \textit{Filter} refers to EKF predict and update steps. Note that, in Table~\ref{tab:compute}, our proposed constraint detection is the only `new' module in what corresponds to a typical perception and localization pipeline for an autonomous vehicle. The local feature matching used in our constraint detection method re-uses local features extracted by HLoc.

\section{Conclusion}

In this paper, we describe a novel and lightweight approach to incorporate limited pose constraint information obtained from detection of dynamic vehicles into a localization system. By detecting and tracking dynamic vehicles over time, we can use the relative position of dynamic vehicles with respect to the autonomous vehicle to add frame-to-frame motion constraints to the pose estimates from PnP-RANSAC. We show that the addition of these constraints yields improved localization compared to an existing PnP+IMU fusion approach almost across all relevant metrics, especially in terms of the operationally critical worst-case localization performance.

There is substantial potential for further work in this area, primarily around building on this concept of using dynamic vehicles as a localization aid. Future work will investigate whether these constraints can be incorporated in a factor graph SLAM algorithm (such as GTSAM~\cite{dellaert2012factor}), which would enable a maximum a posteriori estimation over both measurements and constraints. Future work also includes adding visual odometry to our system, for situations where IMU data is unavailable. Finally, we hope to expand this work to extreme visual appearance conditions (night/rain), sudden velocity variations, and dynamic domains where robots, drones and autonomous systems operate around moving objects, ranging from service robots to drones.

\bibliographystyle{IEEEtran}
\bibliography{references,sg}

\end{document}